\documentclass[conference]{IEEEtran}
\IEEEoverridecommandlockouts
\usepackage{cite}
\usepackage{amsmath,amssymb,amsfonts}
\usepackage{algorithmic}
\usepackage{graphicx}
\usepackage{textcomp}
\usepackage{xcolor}

\usepackage{subfigure}
\usepackage{amsfonts}
\usepackage{xcolor}
\usepackage{amsmath}
\usepackage{makecell}
\usepackage{multirow}
\usepackage{booktabs}

\def\BibTeX{{\rm B\kern-.05em{\sc i\kern-.025em b}\kern-.08em
    T\kern-.1667em\lower.7ex\hbox{E}\kern-.125emX}}
\begin{document}

\title{TPGNN: Learning High-Order Information in Dynamic Graphs via Temporal Propagation}


\author{\IEEEauthorblockN{Zehong Wang}
	\IEEEauthorblockA{\textit{School of Mathematics} \\
		\textit{University of Leeds}\\
		Leeds, United Kingdom \\
		mm22zw@leeds.ac.uk}
	\and
	\IEEEauthorblockN{Qi Li$^*$\thanks{$^*$Corresponding author.}}
	\IEEEauthorblockA{\textit{Department of Computer Science} \\
		\textit{and Engineering, Shaoxing University}\\
		Shaoxing, China \\
		liqi0713@foxmail.com}
	\and
	\IEEEauthorblockN{Donghua Yu}
	\IEEEauthorblockA{\textit{Department of Computer Science} \\
		\textit{and Engineering, Shaoxing University}\\
		Shaoxing, China \\
		donghuayu163@163.com}
}

\maketitle

\begin{abstract}

	Temporal graph is a powerful tool for modeling evolving systems with dynamic interactions. In this paper, we address an important yet neglected problem in temporal graph analysis---\textit{how to efficiently and effectively incorporate information from high-order neighboring nodes}. We identify two challenges in learning high-order information from temporal graphs, i.e., computational inefficiency and over-smoothing, which cannot be solved by conventional techniques applied on static graphs. To address these challenges, we propose a novel temporal propagation-based graph neural network, called TPGNN. Our model consists of two distinct components: propagator and node-wise encoder. The propagator efficiently propagates messages from the anchor node to its temporal neighbors within $k$-hop, and simultaneously updates the state of neighborhoods, achieving efficient computation. The node-wise encoder adopts a transformer architecture to learn node representations by explicitly modeling the importance of memory vectors preserved on the node itself, thus mitigating the over-smoothing. Since the encoding process does not require querying temporal neighbors, our model dramatically reduces time consumption in inference. Extensive experiments on temporal link prediction and node classification demonstrate the superiority of TPGNN over state-of-the-art baselines in terms of efficiency and robustness.

\end{abstract}


\section{Introduction}


Graph serves a fundamental tool for modeling complex systems, representing elements as nodes and their interactions as edges. The task of learning network representations to mine knowledge and discover patterns from graphs has attracted significant research attention across diverse domains, including recommender systems \cite{fan2021continuous}, drug discovery \cite{gilmer2017neural}, traffic prediction \cite{lan2022dstagnn}, and academic networks \cite{wang2022heterogeneous,wang2023sr}. Graph neural networks (GNNs) \cite{kipf2016semi,hamilton2017inductive,velivckovic2017graph} are one of the most influential techniques in graph mining, providing a powerful capacity to jointly model both graph topology and semantics. GNNs have achieved state-of-the-art (SOTA) performance in various downstream tasks, such as node classification, node clustering, and link prediction.


To model the dynamic nature of real-world networks, a significant amount of research has focused on dynamic graphs where each edge is associated with a timestamp \cite{xu2020inductive,rossi2020temporal,wang2020inductive}. For example, in recommender systems, users and items can appear or disappear over time, and attributes on them may also vary across different timestamps. One intuitive approach to represent dynamic graphs is the continuous-time dynamic graph (CTDG) model, where a graph is represented as chronologically arranged edges. Existing algorithms \cite{kumar2019predicting,rossi2020temporal} propose temporal subgraph-aggregation methods to model the dynamics preserved in CTDG. However, to speed up training and inference in downstream tasks, these algorithms only aggregate messages from low-order neighbors (e.g., 1-hop or 2-hop), which fails to capture the knowledge preserved in high-order neighborhoods. Consequently, they inevitably fall short of achieving optimal performance.


In this paper, we address the challenge of \textit{efficiently and effectively learning high-order information in dynamic graphs}. One major obstacle is the \textit{computational inefficiency} in the aggregation process. To better understand this limitation, we analyze the standard CTDG-based algorithm, such as temporal graph attention networks (TGAT) \cite{xu2020inductive}. TGAT consists of two main phases: (1) subgraph generation from a batch of interactions, and (2) recursive graph convolution with time encoding to learn node representations. We argue that the computational inefficiency arises from the aggregation step in graph convolution (i.e., phase 2), which cannot be parallelized for all nodes in the computational graph. Specifically, the message cannot be propagated to low-order nodes until the aggregation on high-order neighbors is complete, resulting in exponential expansion of the computational graph, especially when the model is deep. Moreover, the intermediate state of message passing must be saved, leading to heavy memory consumption.


To address this issue, some methods have proposed directly modeling high-order information using skip-connections or hypergraphs. However, these methods are limited to static graphs and cannot be applied to temporal graphs. For instance, some studies \cite{morris2019weisfeiler,abu2019mixhop} have proposed establishing interactions between the anchor node (i.e., target node and end node in an interaction) and its multi-hop neighbors, to uniformly aggregate messages from $k$-hop neighborhoods in a single layer. However, for dynamic graphs, assigning timestamps to pseudo-links becomes a challenging task. Similarly, some works \cite{feng2019hypergraph} have modeled high-order information using hypergraphs in static networks, but applying this to dynamic graphs would introduce additional computation overhead, which contradicts our original objective. Thus, decomposing the update process and model depth remains a crucial point to ensure efficiency in learning from high-order information in dynamic graphs.


While leveraging high-order information has shown to be useful in GNNs, we still encounter another essential problem---\textit{over-smoothing}. Over-smoothing refers to the phenomenon where increasing the model depth smoothens the node embeddings, rendering nodes indistinguishable and leading to a decline in performance. This problem is contradictory to our intuition that deeper models can provide better expressiveness. Over-smoothing occurs due to the large receptive field of deep models, which covers all nodes in the network. As a result, the learned node representations fail to focus on the low-order neighbors that might provide more discriminative information, leading to similar node embeddings. Consequently, the downstream classifier, which predicts labels based on node embeddings, misclassifies nodes with similar representations yet different labels. Thus, alleviating the over-smoothing problem by emphasizing the importance of low-order neighbors is essential for learning robust and discriminative node representations in dynamic graphs, particularly when incorporating high-order information.


To address the aforementioned deficiencies in CTDG embedding, we propose a novel temporal propagation-based graph neural network (TPGNN). Inspired by APAN \cite{wang2021apan}, TPGNN leverages two key components to learn node representations. The propagator propagates messages from anchor nodes to their $k$-hop neighbors and updates the node state on neighborhoods in parallel. The node-wise encoder incorporates a layer-aware transformer to model the importance of messages from different layers, mitigating the over-smoothing problem by identifying the most influential layer. Additionally, the node-wise encoder aggregates node representations based on node-preserving memories, allowing for decoupling of inference time and model depth. Our extensive experiments on three real-world datasets demonstrate the effectiveness and robustness of TPGNN, showing superior performance compared to SOTA algorithms. We highlight our contributions as follows:
\begin{itemize}
	\item We propose a novel temporal propagation-based graph neural network (TPGNN) to effectively and efficiently learn high-order information in dynamic graphs as well as address the issue of  over-smoothing.
	\item We propose two distinct components, namely propagator and node-wise encoder. The propagator is used to update the state of neighborhoods in different layers in parallel, and the node-wise encoder is designed to aggregate node representation based on node-preserving memories to prevent over-smoothing.
	\item We conduct experiments on three real-world datasets to demonstrate the efficiency and robustness of TPGNN. Extensive experiments also demonstrate that our model is not sensitive to crucial hyper-parameters such as model depth and batch size.
\end{itemize}

\section{Related work}


In recent years, graph representation learning has gained significant attention, particularly for static graphs \cite{perozzi2014deepwalk,grover2016node2vec,tang2015line,kipf2016semi,hamilton2017inductive,velivckovic2017graph,kipf2016variational}. To extend these algorithms to continuously evolving scenarios, researchers have focused on developing temporal graph representation learning methods. One intuitive approach is to split a temporal graph into a sequence of chronological snapshots, and various methods have been proposed to capture the evolving information across all snapshots, including matrix factorization \cite{zhang2018timers}, triadic closure process \cite{zhou2018dynamic}, and random walk \cite{beladev2020tdgraphembed}. Recurrent neural networks (RNNs) \cite{liu2021motif} and transformers \cite{fan2021continuous} have also been utilized to model the chronologically sequential effect across all snapshots to enhance expressiveness. For example, DynGEN \cite{goyal2018dyngem} updates snapshot representations based on the node representations of the previous snapshot, while DySAT \cite{sankar2020dysat} proposes a hierarchical attention mechanism to preserve structural and temporal properties to overcome the time dependency inherent in RNNs.

The previous methods for temporal graph representation learning have mostly focused on discrete approaches, where the temporal information is captured by dividing a temporal graph into a series of snapshots. However, this approach overlooks the dynamics within each snapshot. To address this limitation, some recent works aim to learn the temporal processing of continuously evolving events. For instance, DyRep \cite{trivedi2019dyrep} and its variants \cite{xia2021forecasting,wen2022trend} use temporal Hawkes processes, while CTDNE \cite{nguyen2018continuous}, FiGTNE \cite{liu2020towards}, and CAW-N \cite{wang2020inductive} employ time-respect random walks, time-reinforced random walks, and causal anonymous walks, respectively. GNN-based methods \cite{kumar2019predicting,ma2020streaming,wang2022temporal} leverage time encoding to preserve temporal information. TGAT \cite{xu2020inductive} uses a GAT-like \cite{velivckovic2017graph} architecture with continuous time encoding to encode node representations, TGN \cite{rossi2020temporal} leverages GRU \cite{cho2014learning} to update node representations, and HVGNN \cite{sun2021hyperbolic} extends temporal GNN to hyperbolic space. However, these methods generally aggregate messages from 1-hop or 2-hop neighbors, failing to model high-order information. HIT \cite{xia2021forecasting} proposes to construct temporal hyper-graphs to predict the high-order pattern, introducing extra computational consumption. APAN \cite{wang2021apan} reverses the direction of message passing to reduce inference time. In particular, the messages are propagated from the anchor node to multi-hop neighbors and each node preserves a queue mailbox. Our proposed TPGNN aims to overcome the limitations of previous methods by effectively and efficiently learning high-order information while mitigating over-smoothing.

\begin{figure*}[t]
	\centering
	\includegraphics[width=\linewidth]{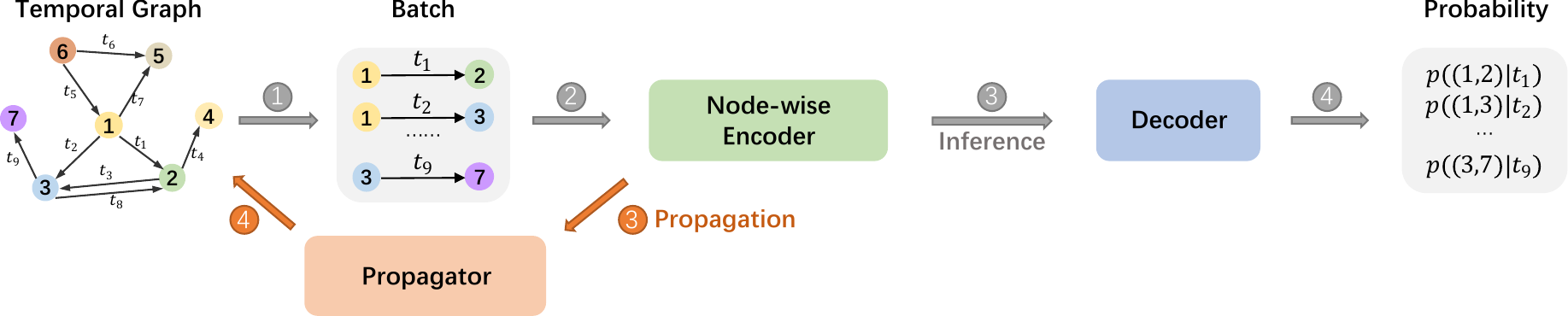}
	\caption{The overview structure of TPGNN. The model comprises two main components: the propagator and the node-wise encoder. First, we sample a batch of interactions from a temporal graph to generate a subgraph, and use the node-wise encoder to learn node representations. Then, we simultaneously feed the learned representations into a decoder for downstream tasks and the propagator to update the node state preserved on multi-hop neighborhoods. This design enables the TPGNN to efficiently learn high-order information while preserving temporal dependencies.}
	\label{fig:framework}
\end{figure*}

\section{Proposed Model}

In this paper, we present the TPGNN model, which consists of the propagator and node-wise encoder, to efficiently gain knowledge from high-order neighbors. The overview architecture of TPGNN is illustrated in Figure \ref{fig:framework}.

\subsection{Preliminary}

A CTDG $\mathcal{G} = \{\mathcal{V}, \mathcal{E}\}$ consists of a node set $\mathcal{V}$ and a time sensitive edge set $\mathcal{E}$. The edge set is represented as a series of time ordered edges: $\mathcal{E} = \{e(0), e(1), ..., e(t)\}$, where $e(t) = (v_s, v_d, e_{sd}, t)$. Note that $v_s \in \mathcal{V}$ and $v_d \in \mathcal{V}$ denote the source and destination nodes, respectively, $e_{sd}$ is the edge feature, and $t$ indicates the timestamp. Our goal is to learn a function to map nodes into embedding space at each timestamp $f: e(t) \to z_s(t), z_d(t)$, where $z_s(t)$ and $z_d(t)$ are embeddings for source and destination nodes at timestamp $t$.

\subsection{Framework}


The proposed TPGNN consists of two main components: the propagator and the node-wise encoder. The propagator is responsible for propagating the knowledge from the anchor nodes to their multi-hop neighbors, updating the node state of neighbors, and injecting high-order information. The node-wise encoder, on the other hand, learns node representations by capturing the importance of messages from multi-hop neighbors, mitigating over-smoothing, and achieving efficiency in inference. In more detail, TPGNN constructs a subgraph by sampling a batch of edges from the temporal graph and aggregates node representations using the node-wise encoder. Then, the learned representations are inputted into the propagator and decoder for updating node states and performing downstream tasks such as link prediction. The model has $k$ layers, and each node in the temporal graph preserves $k$ memories that are updated using messages from the corresponding layer. For instance, the 1-hop memory is updated by the messages from the 1-hop neighbors.

\subsection{Propagator}


The propagator plays a crucial role in our model by propagating messages from the target node to its neighbors along the path of message passing. Compared to existing algorithms such as TGN, our use of the propagator provides three key benefits. First, we update the state of all nodes in the message passing for each interaction, instead of just updating the representation of the target node. This approach preserves more temporality and reduces information loss. Second, since the update process for each layer is independent, we can first propagate messages to neighbors and then parallelize the update process for each neighbor, ensuring efficient computation. Finally, the propagator does not impose significant memory consumption as it does not store intermediate aggregation results.

\begin{figure}[!t]
	\centering
	\includegraphics[width=\linewidth]{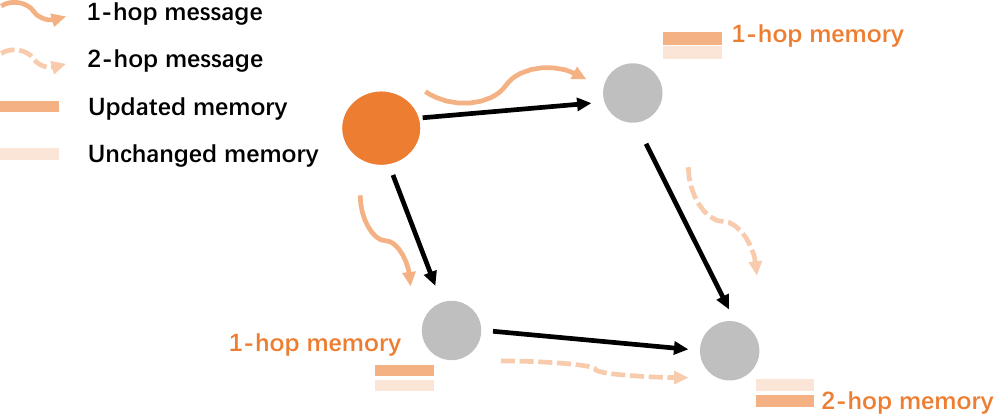}
	\caption{A simplified example of the propagator. The orange node in the graph represents the anchor node, while the grey nodes within $k$-hop distance (in this case, $k=2$) are the influenced nodes. First, messages generated by the anchor node are propagated to its $k$-hop neighbors. Then, the corresponding memory vectors preserved on neighbors are updated simultaneously. It is noteworthy that when multiple messages are propagated to a node, the messages from different sources are combined before updating.}
	\label{fig:propagator}
\end{figure}

Figure \ref{fig:propagator} illustrates a simplified example to demonstrate how the propagator works. In this example, we use orange to indicate the anchor node and grey to denote the influenced nodes. Once the message is generated from the anchor node, we propagate it to its $k$-hop neighbors, where $k=2$ in this example. We then update the memories preserved on the neighbors at each layer concurrently. Specifically, if the influenced node is a direct neighbor (i.e., 1-hop) of the anchor node, we can update its 1-hop memory with the message. Note that when two messages are passed to a node, we first combine them before updating the node's memory.

\subsubsection{Message generator}

In the model, we first generate messages to be propagated based on the newly established interactions in the temporal graph. Specifically, when an interaction $(z_i(t), e_{ij}(t), z_j(t))$ occurs between nodes $v_i$ and $v_j$ at time $t$, where $z_i(t)$ and $z_j(t)$ are the node representations computed by the node-wise encoder, and $e_{ij}(t)$ is the corresponding edge feature, we create separate messages for $v_i$ and $v_j$ as follows:
\begin{align}
	msg_i(t) & = [z_i(t) \Vert e_{ij}(t) \Vert z_j(t)], \\
	msg_j(t) & = [z_j(t) \Vert e_{ij}(t) \Vert z_i(t)],
\end{align}
where $\Vert$ is the concatenation operation. Compared to the summation operation used in APAN \cite{wang2021apan}, concatenation allows for better expressiveness and does not impose any constraints on the node dimension. Note that memory consumption is not a concern in our model, as nodes only need to maintain a small number of memories ($k$) for their $k$-hop neighbors. 

\subsubsection{Message passing}

Once the messages are generated for each anchor node, they are simultaneously propagated to the $k$-hop neighbors. To prevent overload computation, we sample the most recently interacted nodes as temporal neighbors since they are known to preserve essential evolving information \cite{rossi2020temporal}. If a message is propagated to a 2-hop neighbor via a 2-hop path, we refer to it as a 2-hop message, which is then used to update the 2-hop memory. It is noteworthy that a message can be passed to a node via different paths with varying lengths. For simplicity, we leverage the identity function to encode the messages at each layer. We plan to explore more complex message passing functions (e.g., time decay) in future work.

\subsubsection{Message combiner}


When multiple messages are propagated to a node via paths with the same length, we employ a message combiner to fuse the messages. For instance, in Figure \ref{fig:propagator}, where two 2-hop messages (dash line) are simultaneously passed to a node, we first combine these two messages using a fuse function $\theta(\cdot)$, and then use the aggregated message to update the 2-hop memory. We use mean pooling to combine messages instead of summation in practice, as it helps to reduce the impact of nodes with dense connections. The output of the message combiner is a single update message $m_v^n(t)$ for node $v$ in layer $n$ at timestamp $t$.

\begin{figure}[!t]
	\centering
	\includegraphics[width=\linewidth]{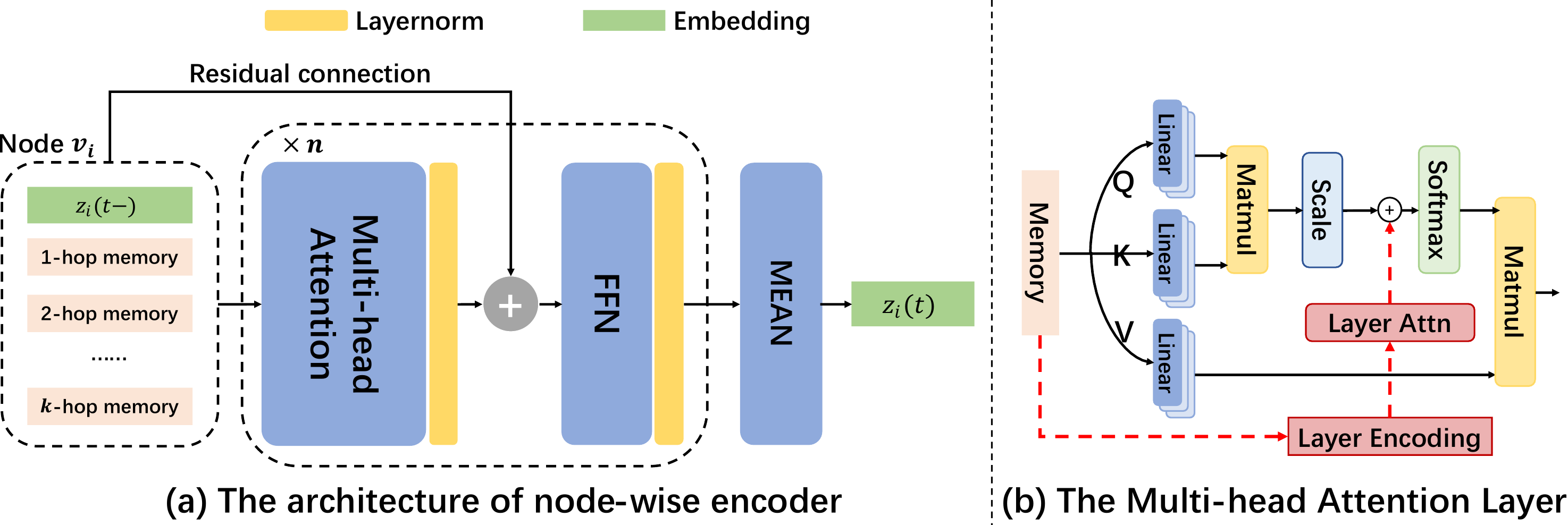}
	\caption{(a) The node-wise encoder aggregates the node state, including node representations and memories, to generate the node representation at the current timestamp using the layer-aware transformer. (b) The transformer contains a multi-head attention and a layer attention, learning the importance of all preserved node memories. }
	\label{fig:node wise encoder}
\end{figure}

\subsubsection{Memory updater}

After combining the messages, we update the memories preserved at each node using a GRU cell due to its ability to maintain temporal properties. We treat the node memory as the hidden state in the GRU and the aggregated message as the input. Note that the updaters for each layer are not parameter-sharing, as the distribution of messages in each layer may significantly fluctuate. By applying the updater, we can simultaneously update all memories preserved by nodes, overcoming the inherent deficiency of existing algorithms to achieve efficient computation. In addition, owing to the messages from a certain layer can only be leveraged to update the corresponding memory vector, we will not mix up the information from different layers, thus mitigating over-smoothing. We formulate the updater as follows:
\begin{align}
	u_v^n(t)               & = \sigma(W_u \cdot m_v^n(t) + U_u \cdot mem_n(t-))          \\
	r_v^n(t)               & = \sigma(W_r \cdot m_v^n(t) + U_r \cdot mem_n(t-))          \\
	\widehat{mem}_v^n(t)   & = W_x \cdot m_v^n(t) + r_v^n(t) \odot U_m \cdot mem_v^n(t-) \\
	\widetilde{mem}_v^n(t) & = (1 - u_v^n(t)) \odot \tanh(\widehat{mem}_v^n(t))          \\
	mem_v^n(t)             & = u_v^n(t) \odot mem_v^n(t-) + \widetilde{mem}_v^n(t)
\end{align}
where $u_v^n(t)$ and $r_v^n(t)$ denote the update gate and reset gate for node $v$ in layer $n$ at timestamp $t$, respectively. $m_v^n(t)$ is the combined message, $mem_v^n(t-)$ indicates the memory vector after the last update, $\sigma(\cdot)$ is the sigmoid function, $\cdot$ and $\odot$ represent dot product and element-wise product, respectively, and $\{W_u, W_r, W_x, U_u, U_r, U_m\}$ are trainable matrix parameters for the GRU cell. We choose GRU over LSTM because it has fewer parameters, ensuring computational efficiency. 

\subsection{Node-wise encoder}


We present the architecture of the proposed node-wise encoder, as shown in Figure \ref{fig:node wise encoder}. This component aims to achieve two main goals: (1) learn the importance of all memories preserved by the node itself, which helps to mitigate over-smoothing; and (2) perform encoding without querying temporal neighborhoods, which significantly reduces the time consumption during inference. Note that the transformer in the node-wise encoder can be stacked in multiple layers to further enhance the model capability and expressiveness.

\subsubsection{Input construction}


To create the input for the node-wise encoder, we combine the node representation learned from the previous update with the multi-hop memories preserved on the nodes. In contrast to using positional encoding, we adopt the identity function to generate the input, as the layer information is already incorporated in the multi-head attention module. This approach avoids introducing excessive inductive bias in the input vectors.

\subsubsection{Multi-head attention}


Once we have constructed the input information, we utilize the multi-head attention layer to infer the temporal representations by weighting the input memories. To address the over-smoothing problem, we also incorporate the layer information in the attention layer as an auxiliary attention. This approach helps the model to learn the relationship between layer and node representations. Intuitively, local neighbors are generally considered to preserve more informative knowledge than high-order neighbors. We initialize the layer embedding $h^{l}_i$ using one-hot encoding, where $i$ is the layer index.

For each node, we use three individual linear transformations to map the input memory at layer $i$ to three vectors, namely query $\mathbf{q}_i(t)$, key $\mathbf{k}_i(t)$, and value $\mathbf{v}_i(t)$. We have the following attention layer to compute the hidden states: 
\begin{equation}
	\begin{aligned}
		Attn & (\mathbf{q}_i(t),\mathbf{k}_i(t),\mathbf{v}_i(t)) =                                                                             \\
		     & Softmax \left(\frac{\mathbf{q}_i(t)^T \cdot \mathbf{k}_i(t)}{\sqrt{d_k}} + a_{layer}^T \cdot h^l_i\right) \times \mathbf{v}(t),
	\end{aligned}
\end{equation}
where $\sqrt{d_k}$ denotes the normalization term, and $a_{layer}$ is an extra attention vector used for computing the layer attention score. To enhance the representativeness of the hidden states, we apply a multi-head mechanism that infers the hidden states of the same node in various semantic spaces. 

\subsubsection{Feed-forward network}


After the multi-head attention layer, we use a feed-forward neural network (FFN) to capture the non-linear relationships between the multi-hop memories. As the range of attention outputs for different nodes may vary, we apply layer normalization (layernorm) to limit the mean and variance, thus preserving the data distribution of a batch. The outputs of FFN are then passed through mean pooling to generate the temporal node representations. By leveraging the node-wise encoder, we can implicitly model the importance of messages from different layers in the neighborhood, overcoming over-smoothing.

\subsection{Discussion}


The proposed TPGNN addresses two essential problems in learning high-order information for dynamic graphs: computational consumption and over-smoothing, which have been neglected in previous works. Our approach is related to the concept of deep graph neural networks \cite{li2019deepgcns,li2020deepergcn,li2021training}, which focuses on building deep GNN models while avoiding heavy computational consumption and over-smoothing. Furthermore, techniques in the line of beyond message passing \cite{wijesinghe2021new,yang2022breaking} can be applied to learn high-order information in graph datasets \cite{morris2019weisfeiler} by extending the Weisfeiler-Leman test. However, none of these methods can be directly generalized to dynamic graphs, as conventional temporal graph algorithms generate subgraphs from a batch of chronological edges, where the number of edges is typically very small (e.g., 200 out of 157,474 interactions for TGN \cite{rossi2020temporal}). 


In addition, we consider that existing algorithms for temporal graphs cannot solve these two limitations. To provide a deeper insight into our declaration, we compare our model with two methods, TGN \cite{rossi2020temporal} and APAN \cite{wang2021apan}. TGN aggregates messages layer by layer, which imposes a bottleneck for computational efficiency and fails to solve the potential over-smoothing. In contrast, our model achieves fast training and inference by decomposing the update process and model depth, as shown in Figure \ref{fig:abl layer}(b)(c). APAN propagates messages from the anchor node to the mailboxes preserved in multi-hop neighborhoods, where messages from neighbors at different layers are mixed, leading to significant over-smoothing. Moreover, APAN drops the oldest message when receiving new information, which exacerbates the over-smoothing, as shown in Figure \ref{fig:abl layer ap}. Although the overview structure of TPGNN is similar to APAN, our model mitigates the over-smoothing by assigning layer-specific memory vectors that can only be updated by messages from neighborhoods located at a certain layer, solving the two inherent limitations of APAN.

\section{Experiments}

\subsection{Experimental Setup}

\subsubsection{Datasets} 


We evaluate the efficiency and robustness of TPGNN against state-of-the-art methods on three public benchmark datasets: Wikipedia, Reddit, and MooC. The statistics of these datasets are presented in Table \ref{tab:dataset}.

\begin{itemize}
	\item \textbf{Wikipedia} is a bipartite temporal graph collected from wikipedia where nodes represent users or wiki-pages and interactions indicate edit operations.
	\item \textbf{Reddit} is a bipartite temporal graph collected from Reddit, where nodes represent users or subreddits, and edges denote post operations.
	\item \textbf{MooC} is a bipartite temporal network containing actions taken by users on the popular MooC platform. Unlike Wikipedia and Reddit, the number of features on each edge in MooC is 4, and the ratio between source nodes and destination nodes is nearly 10:1.
\end{itemize}

\begin{table}[!h]
	\centering
	\resizebox{\linewidth}{!}{
		\begin{tabular}{l | ccc}
			\toprule
			                               & Wikipedia      & Reddit         & MooC           \\ \midrule
			Edges                          & 157,474        & 672,447        & 411,749        \\
			Src nodes                      & 8,227          & 10,000         & 7,047          \\
			Dst nodes                      & 1,000          & 984            & 97             \\
			Edge features                  & 172            & 172            & 4              \\
			Nodes in train.                & 7,475          & 10,844         & 6,625          \\
			Old nodes in val. and test.    & 3,131          & 10,181         & 3,057          \\
			Unseen nodes in val. and test. & 1,752          & 140            & 519            \\
			Data split                     & 70\%-15\%-15\% & 70\%-15\%-15\% & 70\%-15\%-15\% \\ \bottomrule
		\end{tabular}
	}
	\caption{Dataset Statistics.}
	\label{tab:dataset}
\end{table}

\begin{table*}[!t]
	\centering
		\begin{tabular}{l|c|cc|cc|cc}
			\toprule
			\multirow{2}{*}{Methods} & \multirow{2}{*}{Data}                            & \multicolumn{2}{c|}{Wikipedia} & \multicolumn{2}{c|}{Reddit} & \multicolumn{2}{c}{MooC}                                                                                              \\ \cmidrule{3-8}
			                         &                                                  & Accuracy                       & AP                          & Accuracy                    & AP                          & Accuracy                    & AP                          \\ \midrule
			DeepWalk                 & $\mathbf{A}$                                     & 77.32 $\pm$ 0.5                & 90.93 $\pm$ 0.4             & 72.99 $\pm$ 0.4             & 83.02 $\pm$ 0.4             & 59.23 $\pm$ 0.1             & 70.47 $\pm$ 0.1             \\
			Node2vec                 & $\mathbf{A}$                                     & 79.90 $\pm$ 0.7                & 91.89 $\pm$ 0.4             & 73.43 $\pm$ 0.3             & 84.84 $\pm$ 0.4             & 63.20 $\pm$ 0.2             & 76.74 $\pm$ 0.1             \\
			GAE                      & $\mathbf{X}, \mathbf{A}$                         & 73.55 $\pm$ 0.6                & 91.23 $\pm$ 0.4             & 75.95 $\pm$ 0.5             & 93.49 $\pm$ 0.4             & 67.32 $\pm$ 0.2             & 76.72 $\pm$ 0.3             \\
			VGAE                     & $\mathbf{X}, \mathbf{A}$                         & 79.12 $\pm$ 0.2                & 91.75 $\pm$ 0.5             & 75.89 $\pm$ 0.2             & 92.29 $\pm$ 0.2             & 67.72 $\pm$ 0.2             & 77.14 $\pm$ 0.2             \\
			GAT                      & $\mathbf{X}, \mathbf{A}, \mathbf{Y}$             & 89.43 $\pm$ 0.3                & 94.01 $\pm$ 0.6             & 92.51 $\pm$ 0.5             & 97.03 $\pm$ 0.3             & 92.67 $\pm$ 0.2             & 92.57 $\pm$ 0.1             \\
			SAGE                     & $\mathbf{X}, \mathbf{A}, \mathbf{Y}$             & 87.85 $\pm$ 0.4                & 93.16 $\pm$ 0.3             & 92.84 $\pm$ 0.6             & 97.55 $\pm$ 0.4             & 81.23 $\pm$ 0.2             & 92.31 $\pm$ 0.2             \\ \midrule
			CTDNE                    & $\mathbf{X}, \mathbf{A}, \mathbf{T}$             & 82.17 $\pm$ 0.4                & 92.39 $\pm$ 0.5             & 77.91 $\pm$ 0.4             & 91.53 $\pm$ 0.3             & 79.22 $\pm$ 0.3             & 83.86 $\pm$ 0.5             \\
			DyRep                    & $\mathbf{X}, \mathbf{A}, \mathbf{Y}, \mathbf{T}$ & 90.77 $\pm$ 0.3                & 94.53 $\pm$ 0.2             & 93.37 $\pm$ 0.3             & 98.01 $\pm$ 0.3             & 79.80 $\pm$ 0.2             & 75.22 $\pm$ 0.1             \\
			JODIE                    & $\mathbf{X}, \mathbf{A}, \mathbf{Y}, \mathbf{T}$ & 90.83 $\pm$ 0.6                & 94.10 $\pm$ 0.5             & 92.77 $\pm$ 0.4             & 97.40 $\pm$ 0.5             & 77.63 $\pm$ 0.7             & 79.23 $\pm$ 1.2             \\
			TGAT                     & $\mathbf{X}, \mathbf{A}, \mathbf{Y}, \mathbf{T}$ & 91.14 $\pm$ 0,3                & 96.55 $\pm$ 0.1             & 93.92 $\pm$ 0.3             & 98.23 $\pm$ 0.3             & 70.62 $\pm$ 0.3             & 76.32 $\pm$ 0.5             \\
			TGN                      & $\mathbf{X}, \mathbf{A}, \mathbf{Y}, \mathbf{T}$ & 93.32 $\pm$ 0.3                & \underline{98.45 $\pm$ 0.1} & \underline{95.38} $\pm$ 0.3 & \underline{98.80 $\pm$ 0.1} & 83.64 $\pm$ 0.9             & 89.29 $\pm$ 1.0             \\
			HVGNN                    & $\mathbf{X}, \mathbf{A}, \mathbf{Y}, \mathbf{T}$ & \underline{93.44 $\pm$ 0.4}    & 98.02 $\pm$ 0.1             & 94.22 $\pm$ 0.4             & 98.59 $\pm$ 0.4             & 84.54 $\pm$ 0.4             & 88.83 $\pm$ 0.3             \\
			APAN                     & $\mathbf{X}, \mathbf{A}, \mathbf{Y}, \mathbf{T}$ & 92.97 $\pm$ 0.3                & 98.24 $\pm$ 0.2             & 95.02 $\pm$ 0.2             & 98.65 $\pm$ 0.3             & \underline{98.60 $\pm$ 0.4} & \underline{98.71 $\pm$ 0.2} \\ \midrule
			\textbf{TPGNN w/o LA}    & $\mathbf{X}, \mathbf{A}, \mathbf{Y}, \mathbf{T}$ & 95.07 $\pm$ 0.2                & 98.54 $\pm$ 0.1             & 97.02 $\pm$ 0.1             & 98.88 $\pm$ 0.2             & 98.98 $\pm$ 0.4             & 99.04 $\pm$ 0.5             \\
			\textbf{TPGNN}           & $\mathbf{X}, \mathbf{A}, \mathbf{Y}, \mathbf{T}$ & \textbf{95.47 $\pm$ 0.3}       & \textbf{98.82 $\pm$ 0.1}    & \textbf{97.31 $\pm$ 0.1}    & \textbf{99.49 $\pm$ 0.1}    & \textbf{99.05 $\pm$ 0.5}    & \textbf{99.13 $\pm$ 0.5}    \\ \bottomrule
		\end{tabular}

	\caption{Temporal link prediction results on three benchmark datasets in terms of Accuracy (\%) and average precision (AP) (\%). Boldface and underline denote the best performance of the variants of TPGNN and other baselines, respectively. $\mathbf{T}$ denotes the preservation of temporal information and \textbf{LA} indicate the layer attention in the node-wise encoder. }
	\label{tab:link prediction}
\end{table*}

\subsubsection{Baselines}

We compare TPGNN with various baselines, ranging from static to dynamic graph embedding models.

\begin{itemize}
	\item DeepWalk \cite{perozzi2014deepwalk} and Node2vec \cite{grover2016node2vec} apply random walk to generate node sequence and leverage the skip-gram model to learn node representations.
	\item GAE \cite{kipf2016variational} and VGAE \cite{kipf2016variational} leverage GCN \cite{kipf2016semi} as the encoder and perform graph reconstruction as the task to compute the loss.
	\item GAT \cite{velivckovic2017graph} leverages the attention layer to assign the importance of each node and corresponding neighborhoods.
	\item GraphSAGE (SAGE) \cite{hamilton2017inductive} performs graph sampling to learn embeddings of unseen nodes.
	\item CTDNE \cite{nguyen2018dynamic} is the extension of DeepWalk in the dynamic setting.
	\item DyRep \cite{trivedi2019dyrep} applies Hawkes process to model the interaction generation process in temporal graphs.
	\item JODIE \cite{kumar2019predicting} proposes to aggregate the temporal information from 1-hop neighbors for nodes in temporal graphs.
	\item TGAT \cite{xu2020inductive} extends GraphSAGE on temporal graphs by introducing time encoding in aggregation.
	\item TGN \cite{rossi2020temporal} applies the time-aware updater and aggregator to achieve SOTA performance but imposes a large computational cost.
	\item HVGNN \cite{sun2021hyperbolic} applies aggregation in hyperbolic space to enhance expressiveness.
	\item APAN \cite{wang2021apan} proposes to use the propagator and mailbox-based aggregator to speed up the inference, whereas failing to mitigate over-smoothing.
\end{itemize}

\begin{table}[!t]
	\centering
		\begin{tabular}{l|cc}
			\toprule
			Methods               & Wikipedia                   & Reddit                      \\ \midrule
			GAT                   & 81.50 $\pm$ 0.9             & 63.59 $\pm$ 0.7             \\
			SAGE                  & 81.39 $\pm$ 0.7             & 60.45 $\pm$ 0.4             \\
			CTDNE                 & 77.01 $\pm$ 0.3             & 61.06 $\pm$ 0.7             \\
			DyRep                 & 84.66 $\pm$ 0.7             & 64.31 $\pm$ 1.1             \\
			JODIE                 & 83.93 $\pm$ 0.3             & 60.97 $\pm$ 0.9             \\
			TGAT                  & 84.00 $\pm$ 0.4             & 66.85 $\pm$ 0.9             \\
			TGN                   & 87.12 $\pm$ 0.4             & \underline{68.41 $\pm$ 0.6} \\
			HVGNN                 & 87.90 $\pm$ 0.3             & 68.09 $\pm$ 0.5             \\
			APAN                  & \underline{88.86 $\pm$ 0.3} & 66.39 $\pm$ 0.4             \\ \midrule
			\textbf{TPGNN w/o LA} & 89.39 $\pm$ 0.3             & 70.41 $\pm$ 0.5             \\
			\textbf{TPGNN}        & \textbf{89.65 $\pm$ 0.4}    & \textbf{70.83 $\pm$ 0.6}    \\ \bottomrule
		\end{tabular}

	\caption{Node classification with ROC-AUC (\%).}
	\label{tab:node classification}
\end{table}

\subsubsection{Evaluation protocol}

We evaluate the performance of TPGNN on two tasks: temporal link prediction and node classification. To mitigate randomness, we perform 10 runs of each baseline and report their average results. For temporal link prediction, we follow the protocol \cite{wang2021apan}, which focuses on predicting the existence of links given previous interactions. We measure the accuracy and average precision (AP) to evaluate the performance. To sample negative interactions based on the current timestamp in the transductive setting, we use a chronological batch sampling strategy and a corresponding negative sampling strategy \cite{xu2020inductive,wang2021apan}. We employ a 2-layer MLP as the decoder to predict the probabilities of interaction generation, using the concatenation of node representations and edge embeddings, denoted as $edge_{i,j,t} = (z_i(t) \Vert e_{ij}(t) \Vert z_j(t))$. We use cross-entropy loss to train the model:
\begin{equation}
	\begin{aligned}
		\mathcal{L} = \sum_{(v_i, v_j, e_{ij}, t) \in \mathcal{G}} & -log(\sigma(MLP(edge_{i,j,t})))                               \\
		                                                           & - \mathbb{E}_{v_n \sim P_n(v)}log(\sigma(MLP(edge_{i,n,t}))),
	\end{aligned}
\end{equation}
where $P_n(v)$ is the negative sampling distribution, $\sigma(\cdot)$ is the sigmoid function, and $MLP(\cdot)$ is the decoder. For node classification, we also adopt a 2-layer MLP following a $Softmax(\cdot)$ as the decoder to predict the label of the node based on the current node representation $z(t)$. We use AUC as the evaluation metric.

\subsubsection{Implementation details}

For all baselines, we used Xavier initialization to initialize the parameters and Adam as the optimizer. We perform a grid search to tune the hyperparameters, including the learning rate from $1e-5$ to $1e-3$, the number of neighbors in $\{10, 15, 20, 25, 30\}$, the dropout rate from $0.1$ to $0.3$, the batch size as $200$, early stopping as $5$, node dimension as $172$, and the number of message passing layers as $2$. For our proposed model, we set the learning rate to $1e-4$, the number of neighbors to $20$, the dropout rate to $0.1$, and the number of layers to $5$. In addition, we set the number of attention heads to $2$ for attention-based methods, and the transformer layers to $1$ for transformer-based models. For the decoders, we use the $ReLU(\cdot) = max(\cdot, 0)$ as the activation of the first layer and the sigmoid $\sigma(\cdot)$ for the last layer to predict probabilities. For TGN and APAN, we follow the hyper-parameters reported in the original paper \cite{rossi2020temporal,wang2021apan}.

\begin{figure*}[!t]
	\centering
	\subfigure[Average Precision]{
		\includegraphics[width=0.3\linewidth]{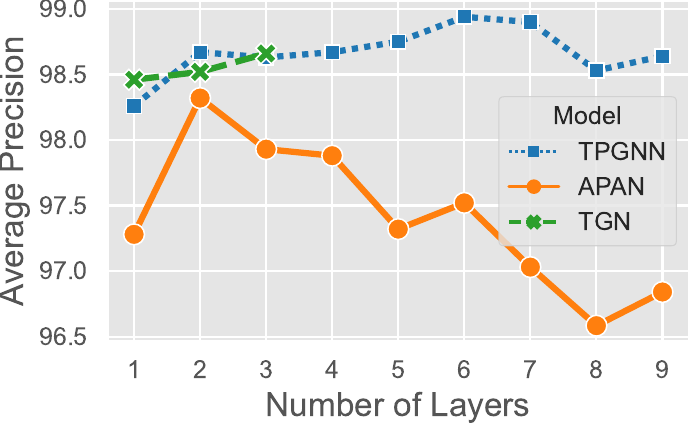}
		\label{fig:abl layer ap}
	}
	\subfigure[Training Time]{
		\includegraphics[width=0.3\linewidth]{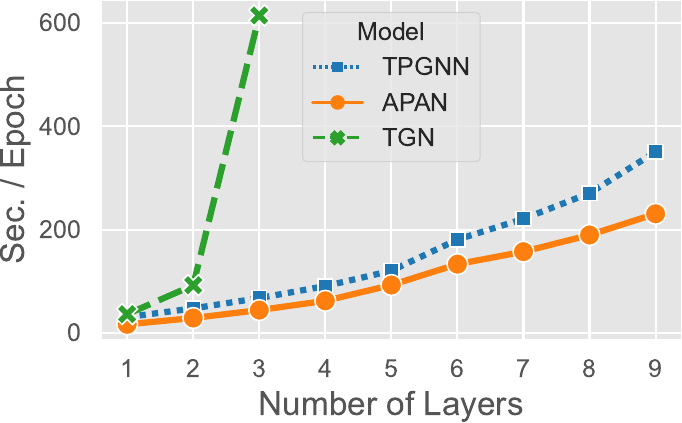}
		\label{fig:abl layer train}
	}
	\subfigure[Inference Time]{
		\includegraphics[width=0.3\linewidth]{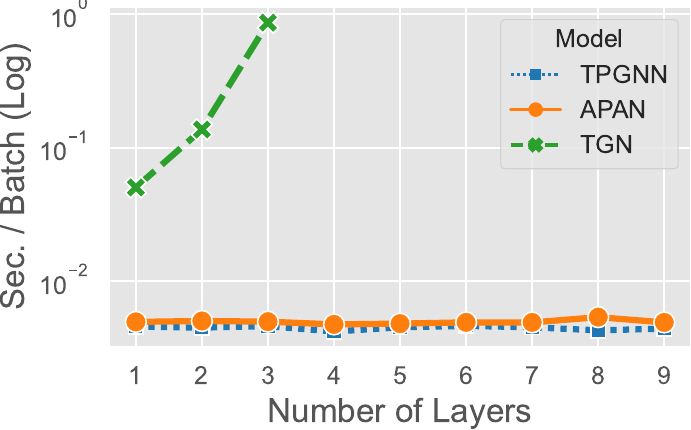}
		\label{fig:abl layer infer}
	}
	\caption{The average precision (AP), training time per epoch, and inference time per batch in terms of the number of layers on Wikipedia. Our model (TPGNN) is capable of flexible extension to multiple layers while maintaining high efficiency in both training and inference, as well as overcoming over-smoothing. It is worth noting that TGN encounters out-of-memory (OOM) issues after three layers in a 12GB TITAN V.}
	\label{fig:abl layer}
\end{figure*}

\begin{figure*}[!t]
	\centering
	\subfigure[Batch Size on Wikipedia]{
		\includegraphics[width=0.3\linewidth]{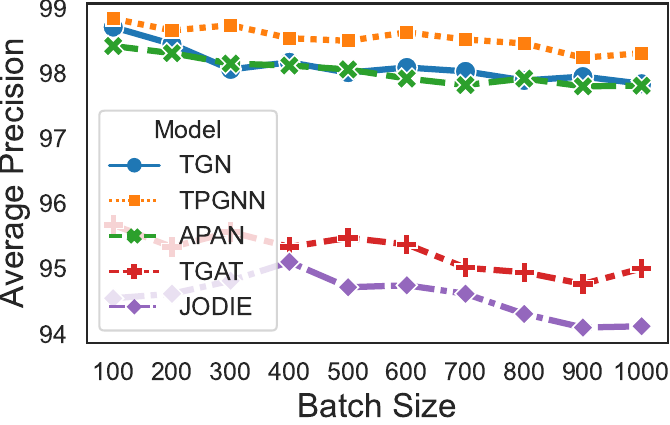}
		\label{fig:abl bs}
	}
	\subfigure[Number of Neighbors]{
		\includegraphics[width=0.3\linewidth]{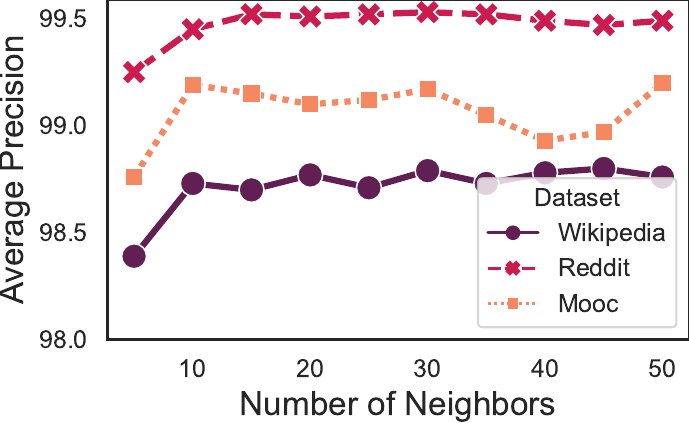}
		\label{fig:abl neighbor}
	}
	\subfigure[Node Dimension]{
		\includegraphics[width=0.3\linewidth]{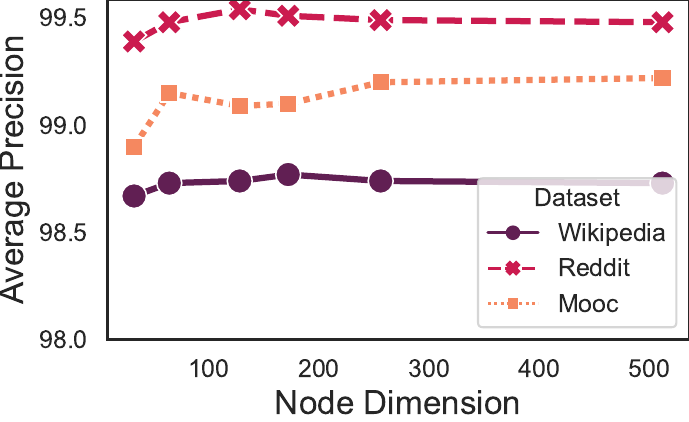}
		\label{fig:abl dim}
	}
	\caption{Hyperparameter analysis on link prediction in terms of (a) batch size on Wikipedia, (b) number of neighbors, and (c) node dimension.}
	\label{fig:abl remaining}
\end{figure*}

\subsection{Quantitative Results}


The results of our experiments on temporal link prediction and node classification are presented in Table \ref{tab:link prediction} and \ref{tab:node classification}, respectively. We observe that dynamic models consistently outperform static models across all datasets, indicating the importance of preserving temporal information. Our proposed TPGNN outperforms state-of-the-art algorithms on all datasets, demonstrating the effectiveness of learning high-order information. Incorporating layer information in the node-wise encoder results in more informative node representations. Even when the best parameters for link prediction are used for node classification, our model still achieves the best performance, highlighting the significance of learning high-order information for improving discrimination. APAN and TPGNN significantly outperform other dynamic methods on the MooC dataset, which is possibly due to the ability of these methods to update the state of multi-hop neighbors when an interaction is established, overcoming the dynamic loss. In contrast, typical dynamic methods such as TGAT and TGN only update the state of the target nodes. 

\subsection{Hyperparameter Sensitivity}

\subsubsection{Number of layers}


In Figure \ref{fig:abl layer}, we compare the efficiency of TPGNN with TGN and APAN when learning high-order information, by varying the number of layers in the models. We observe that TPGNN and APAN can be extended to multiple layers without imposing heavy time consumption in training and inference. However, for TGN, the time consumption increases exponentially in both training and inference, and the model suffers from out-of-memory (OOM) when the depth is greater than 3. This is because TPGNN and APAN concurrently update the node state preserved by multi-hop neighbors via propagator, and learn the node representation by node-level aggregation without querying temporal neighbors. Furthermore, due to over-smoothing, the performance of APAN decreases significantly with the growth of layers, while the performance of our model fluctuates but still improves, showing the necessity of individually leveraging messages from different layers to solve over-smoothing. We believe the model degradation of APAN is due to the inherent drawback of the mailbox, where messages from different layers are chaotically mixed up.

\subsubsection{Batch size}


We conduct a sensitivity analysis on batch size to evaluate the performance of existing CTDG-based models in preserving temporal information. As shown in Figure \ref{fig:abl bs}, we found that our model is less sensitive to changes in batch size compared to SOTA baselines. This is likely due to the propagation mechanism that updates the state of nodes within the message passing, which helps to resist information loss. In contrast, existing methods such as TGN only update the state of target nodes, which may result in a more severe information loss as the batch size increases. We speculate that our model capacity to preserve information may be due to this difference in propagation mechanism.

\subsubsection{Number of neighbors}


In Figure \ref{fig:abl neighbor}, we present the impact of the number of sampled neighbors on the performance of TPGNN across three datasets. Our results show that the model performs the best when the number of neighbors is in the median range. If the number of sampled neighbors is too small or too large, the model performance decays. We hypothesize that too few neighbors may result in important nodes being missed, while too many neighbors may contain duplicates. Despite the impact of the hyperparameter on the model performance, our model is observed to be robust and not sensitive to the number of neighbors, indicating its robustness.

\subsubsection{Node dimension}


In Figure \ref{fig:abl dim}, we examine the impact of the node dimension on performance. The figure demonstrates that the performance of the model remains relatively stable under different node dimensions. This observation indicates that TPGNN has the potential to be deployed in the real world since it requires low memory consumption with a small node dimension.

\section{Conclusion}


In this paper, we proposed TPGNN, a temporal propagation-based graph neural network, to address the issues of computational inefficiency and over-smoothing in learning high-order information for temporal graphs. Our model consists of two main components, the propagator and node-wise encoder, which propagate messages efficiently and aggregate node representations based on states preserved on the node itself to mitigate over-smoothing. By not requiring the querying of temporal neighbors during inference, we further speed up the process. Experimental results on temporal link prediction and node classification show that TPGNN outperforms state-of-the-art methods. Additionally, extensive experiments demonstrate the robustness of our model on significant hyper-parameters, including model depth and batch size.

\section{Acknowledgement}

This work was supported in part by the National Natural Science Foundation of China under Grant No. 62002227 and No. 62002226, and the Zhejiang Natural Science Foundation of China under No. LY22F020003.

\bibliographystyle{IEEEtran}
\bibliography{tpgnn}

\end{document}